\documentclass[letterpaper, 10 pt, journal, twoside]{IEEEtran} 
\IEEEoverridecommandlockouts 

\usepackage{hyperref, url, amsmath, amssymb, booktabs, graphicx, color}
\graphicspath{{images/}}

\title{Classification of Household Materials via Spectroscopy} 

\author{Zackory Erickson$^{1}$, Nathan Luskey$^{1}$, Sonia Chernova$^{2}$, and Charles C. Kemp$^{1}$%
\thanks{Manuscript received: September, 10, 2018; Revised November, 30, 2018; Accepted December, 26, 2018}
\thanks{This paper was recommended for publication by Editor Han Ding upon evaluation of the Associate Editor and Reviewers' comments. This work was supported by NSF award IIS-1514258, by NIDILRR, grant 90RE5016-01-00 via RERC TechSAge, by a Google Faculty Research Award, and by AWS Cloud Credits for Research.}
\thanks{$^{1}$Zackory Erickson, Nathan Luskey, and Charles C. Kemp are with the Healthcare Robotics Lab, Georgia Institute of Technology, Atlanta, GA., USA.
{\tt\small zackory@gatech.edu}; {\tt\small nathanluskey@gatech.edu}; {\tt\small charlie.kemp@bme.gatech.edu}}%
\thanks{$^{2}$Sonia Chernova is with the Robot Autonomy and Interactive Learning Lab, Georgia Institute of Technology, Atlanta, GA., USA.
{\tt\small chernova@cc.gatech.edu}}%
\thanks{Digital Object Identifier (DOI): see top of this page.} 
}

\begin{document}

\maketitle

\begin{abstract}

Recognizing an object's material can inform a robot on the object's fragility or appropriate use. To estimate an object's material during manipulation, many prior works have explored the use of haptic sensing. In this paper, we explore a technique for robots to estimate the materials of objects using spectroscopy. We demonstrate that spectrometers provide several benefits for material recognition, including fast response times and accurate measurements with low noise. Furthermore, spectrometers do not require direct contact with an object. To explore this, we collected a dataset of spectral measurements from two commercially available spectrometers during which a robotic platform interacted with 50 flat material objects, and we show that a neural network model can accurately analyze these measurements. Due to the similarity between consecutive spectral measurements, our model achieved a material classification accuracy of 94.6\% when given only one spectral sample per object. Similar to prior works with haptic sensors, we found that generalizing material recognition to new objects posed a greater challenge, for which we achieved an accuracy of 79.1\% via leave-one-object-out cross-validation. Finally, we demonstrate how a PR2 robot can leverage spectrometers to estimate the materials of everyday objects found in the home. From this work, we find that spectroscopy poses a promising approach for material classification during robotic manipulation.
\end{abstract}

\begin{IEEEkeywords}
Perception for Grasping and Manipulation, Haptics and Haptic Interfaces, Mobile Manipulation
\end{IEEEkeywords}

\section{INTRODUCTION}
\label{sec:intro}

\IEEEPARstart{T}{he} materials that form an object have important implications as robots interact with people and manipulate objects in real-world environments. 
For example, the fragility, hardness, safety, and flexibility of an object depend on its materials. By identifying household materials, robots could behave more intelligently when working with household objects. A robot might choose an object based on its material, such as a plastic container for the microwave; handle a fragile object more carefully, such as avoiding creasing or tearing paper; or keep a sharp metal object away from a person.
To achieve this form of material recognition, several prior works have replicated properties of the human somatosensory system, such as force, temperature, or vibration sensing~\cite{decherchi2011tactile, bhattacharjee2015material, jamali2010material}.

\begin{figure}
\centering
\includegraphics[width=0.45\textwidth, trim={0cm 5cm 0cm 0cm}, clip]{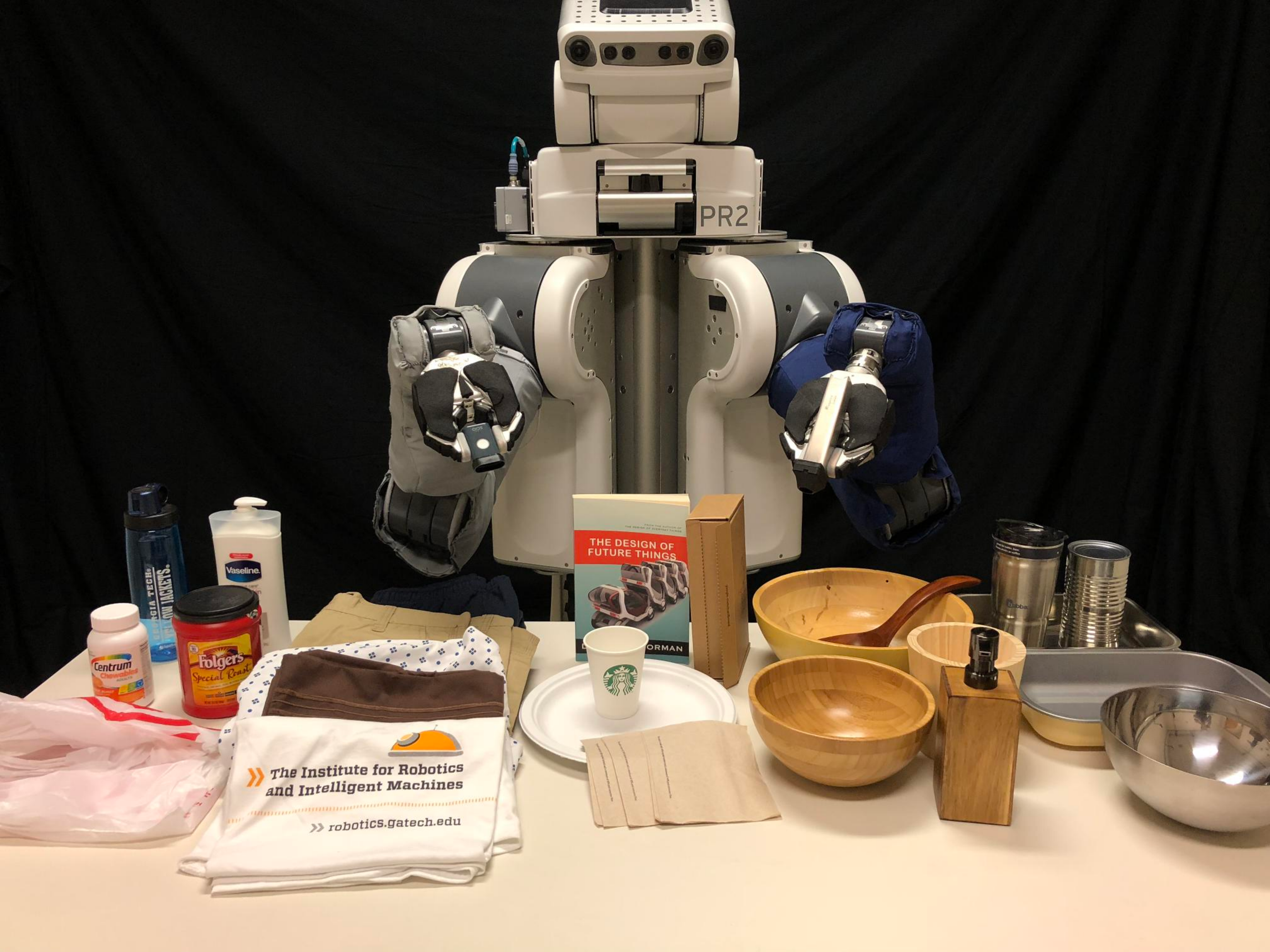}
\caption{\label{fig:pr2_objects}The PR2 used both spectrometers, independently, to estimate the materials of 25 everyday objects. These objects span five material categories: plastic, fabric, paper, wood, and metal.}
\vspace{-0.4cm}
\end{figure}

In this paper, we explore a technique for robots to estimate the materials of objects through the use of spectroscopy. The study of spectroscopy involves measuring the interaction between electromagnetic radiation and matter. Informally, this is the process of measuring the intensity of light reflected back from an object as a function of the light's wavelength. Spectrometers (or spectroscopy sensors) present several benefits for material recognition on a robot, including fast sensing capabilities with accurate and repeatable measurements. In addition, spectrometers do not require direct physical contact with an object, unlike many haptic sensors.



\begin{figure}
\centering
\vspace{6pt}
\includegraphics[width=0.48\textwidth, trim={0cm 0cm 0cm 0cm}, clip]{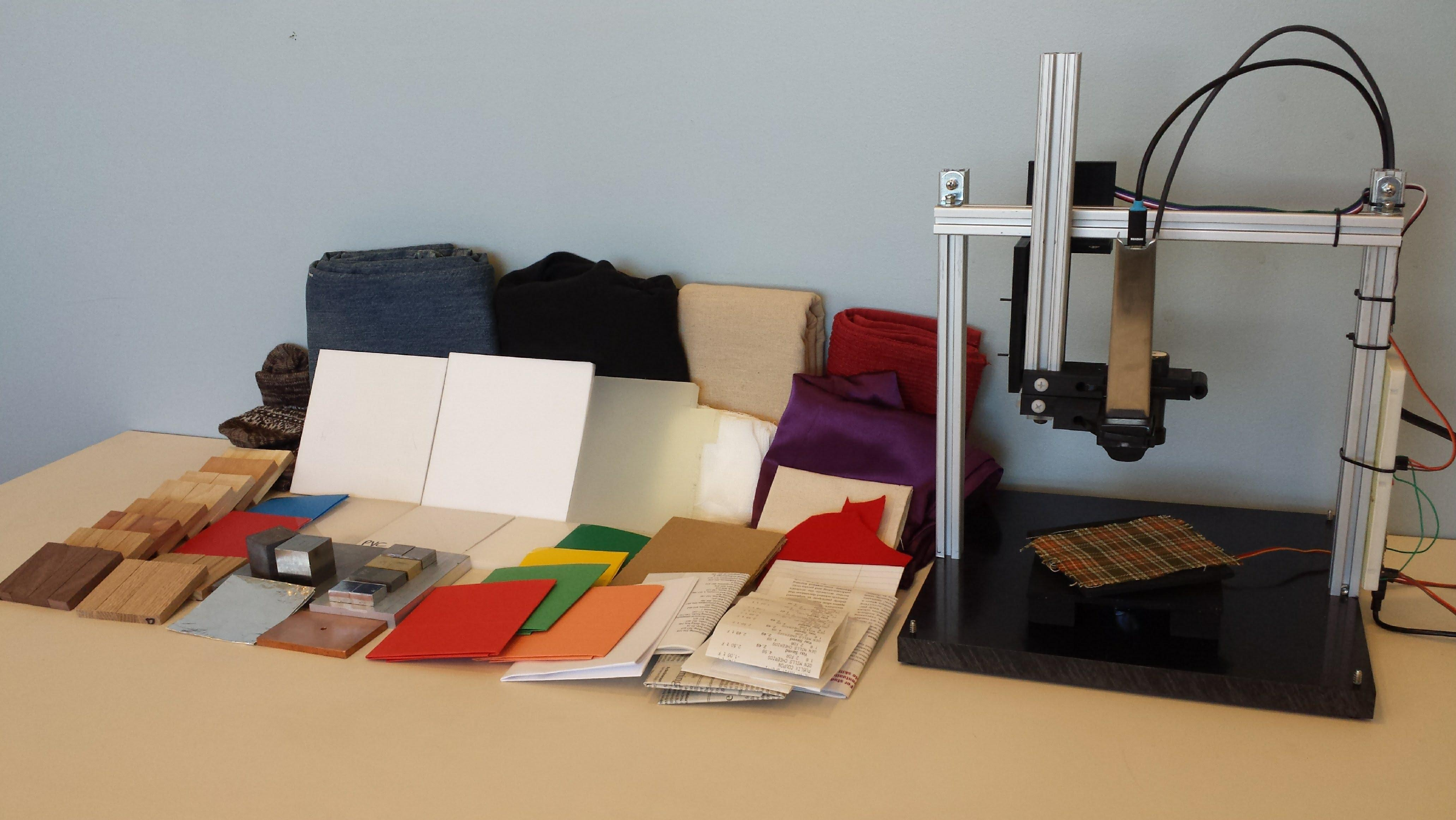}
\caption{\label{fig:intro}Robotic platform and two micro spectrometers along with all 50 objects from the five material categories.}
\vspace{-0.4cm}
\end{figure}

We evaluate and compare two commercially available handheld micro spectrometers for material inference, seen in Fig.~\ref{fig:pr2_objects}, each of which covers a different region of the electromagnetic spectrum. Together, these sensors measure wavelengths ranging between $\lambda=$~317~nm and $\lambda=$~1,070~nm, which spans the near-ultraviolet, visible, and near-infrared light spectra. As part of this study, we have released a new dataset\footnote{Dataset: \url{https://pwp.gatech.edu/hrl/smm50/}} consisting of spectral measurements collected from an idealized robotic platform that interacted with 50 objects from five material categories (metal, plastic, wood, paper, and fabric), as seen in Fig.~\ref{fig:intro}. With spectral data, we achieved state-of-the-art material classification performance of 99.9\% accuracy over 5-fold cross-validation. 
In addition, we show how a PR2, mobile manipulator, can use spectroscopy to accurately estimate the materials of several household objects.

\pagebreak
In this paper, we make the following contributions:
\begin{itemize}
  \item We demonstrate how robots can leverage spectral data to estimate the material of an object.
  \item We compare two commercially available micro spectrometers for material recognition, which both operate at different regions of the electromagnetic spectrum.
  \item We present and analyze a new dataset of 10,000 spectral measurements collected from 50 different objects across five material categories.
  \item We demonstrate how a mobile manipulator, PR2 robot, can leverage near-infrared spectroscopy to estimate the materials of everyday objects prior to manipulation.
\end{itemize}

Our results demonstrate that spectroscopy serves as a viable technique to material recognition, with competitive performance to existing haptic sensing approaches, quick sensing times, low noise, and robustness to changes in both object shapes and robotic platforms.

\section{RELATED WORK}
\label{sec:related_work}


\subsection{Material/Haptic Recognition}


Many prior studies have successfully demonstrated that robots can leverage haptic sensing to estimate material properties of objects. In comparison to these approaches, spectroscopy offers many benefits, including fast response times and highly repeatable samples with little noise between consecutive measurements of the same object. 

The BioTac is a commonly used sensor for haptic perception that is capable of sensing force, temperature, and vibration~\cite{fishel2012bayesian}. For example, Kerr et al. collected haptic features from a BioTac sensor while a robot performed pressing and sliding interactions with objects from six material categories~\cite{kerr2018material}. With an SVM model, the researchers achieved a classification accuracy of 86.19\% given 20 seconds of contact per trial.

In a similar study to ours, Bhattacharjee et al. presented the use of active temperature sensing to estimate the materials of objects given short durations of contact~\cite{bhattacharjee2015material}. The authors used a 1-DoF robot to make contact with 11 flat material samples. By training an SVM on 500 interactions per material, they achieved a material classification accuracy of 98\% with 1.5 seconds of contact and 84\% with 0.5 seconds of temperature data per interaction. We also used a 1-DoF robot to collect spectral samples from flat objects, yet we further evaluated how a PR2 robot can leverage spectroscopy to estimate the materials of common household objects. When interacting with everyday objects, spectroscopy offers the flexibility of not requiring physical contact with an object, which allows a robot to estimate material properties before manipulation.


Sinapov et al. used a dual arm humanoid robot to recognize the semantic labels of 100 objects across 20 object categories, such as tin cans, pasta boxes, and bottles~\cite{sinapov2014grounding}. Their robot performed 10 exploratory behaviors with each object, while measuring vision, proprioception, and audio sensory modalities. Sinapov et al. then proposed a technique for recognizing 20 surfaces, such as paper, wood, cotton, and leather~\cite{sinapov2011vibrotactile}. The robot performed five exploratory behaviors ranging in durations of 3.9 to 14.7 seconds with each surface. The authors achieved a classification accuracy of 80.0\% using an SVM with 10-fold cross-validation. Unlike exploratory behavior approaches, spectroscopy presents an opportunity to quickly estimate material properties of an object without direct contact. 

Several works have also used visual features to estimate the material label of an object~\cite{liu2010exploring, wang2016dataset}. For example, Bell et al. fine-tuned a convolutional neural network (CNN) model for pixel level material classification of images~\cite{bell15minc}. Gao et al. coupled this visual CNN model with a CNN trained on haptic features from a BioTac to estimate haptic adjectives (e.g. fuzzy or soft)~\cite{gao2016deep}. Measuring an object's surface reflectance, or bidirectional reflectance distribution function (BRDF), can also help estimate material properties~\cite{liu2014discriminative} and construct accurate material visualizations for simulated objects~\cite{matusik2003data}.

In prior work, we designed a semi-supervised learning approach to reduce the amount of labeled haptic data needed for accurate material recognition with a robot~\cite{erickson2017semi}. A PR2 performed nonprehensile interactions with 72 objects from 6 material categories, while measuring force, temperature, and vibration haptic signals. With a generative adversarial network (GAN), the robot was able to correctly classify the materials of household objects with $\sim$90\% accuracy when 92\% of the training data were unlabeled. In comparison, this paper focuses on supervised learning and explores a sensory modality---spectroscopy---that is unique from the haptic modalities found in the human somatosensory system. 


\subsection{Spectroscopy}

Types of spectroscopy often differ based on the type of electromagnetic radiation involved in the interaction, where each type can be classified by an associated wavelength region of the spectrum. In this work, we evaluate spectroscopy for material classification using the near-ultraviolet, visible, and near-infrared spectra.


Near-infrared spectroscopy (with wavelengths varying from 700 to 2,500~nm) has been widely used for drug analysis and quality control within the pharmaceutical industry~\cite{reich2005near, roggo2007review}. Near-infrared spectroscopy has also become the standard for analyzing biological components in food manufacturing, as discussed by Williams and Norris~\cite{williams1987near}. As an example of this, both visible and near-infrared spectroscopy have been used for tea and food quality identification~\cite{he2007discrimination, cen2007theory}. He et al. trained a neural network to recognize eight tea varieties given 200 total spectral samples, and the authors achieved 100\% classification accuracy when testing on 40 new samples~\cite{he2007discrimination}. Near-infrared spectroscopy has also been applied to the identification of polymer resins. Masoumi et al. presented an approach to separate different recyclable plastics using spectral measurements between the 700-2,000~nm range~\cite{masoumi2012identification}.

Recently, a number of commercial handheld near-infrared spectrometers have been released~\cite{rateni2017smartphone}. For example, Das et al. introduced a handheld, 3D printed spectrometer that measured wavelengths in the range of 340-780~nm and was used to test fruit ripeness~\cite{das2016ultra}. 
SCiO, another handheld spectrometer, is commercially available and is commonly promoted for its use in food analysis and estimating the freshness of produce~\cite{lee2017nir}. 
Handheld sensors have also been applied to object recognition, where Yeo et al. used the front-facing camera of a smartphone to capture 30 spectral measurements each from 30 household objects~\cite{yeo2017specam}. The authors then trained an SVM to estimate the object label for a given spectral measurement and they achieved an accuracy of 99.44\% using 10-fold cross validation. Similarly, we collect spectral measurements from handheld sensors; however, we focus our work on general material category classification and we evaluate how well spectroscopy generalizes to estimating the materials of entirely new objects.

\section{SPECTROSCOPY DATASET}
\label{sec:spectrometer}

In order to evaluate the use of spectroscopy for material recognition, we collected a dataset of spectral measurements taken from two different spectrometers attached to a 1 degree of freedom (DoF) robot. 

\subsection{Spectrometers}

\begin{figure}
\centering
\vspace{6pt}
\includegraphics[width=0.23\textwidth, trim={5cm 1cm 3cm 3cm}, clip]{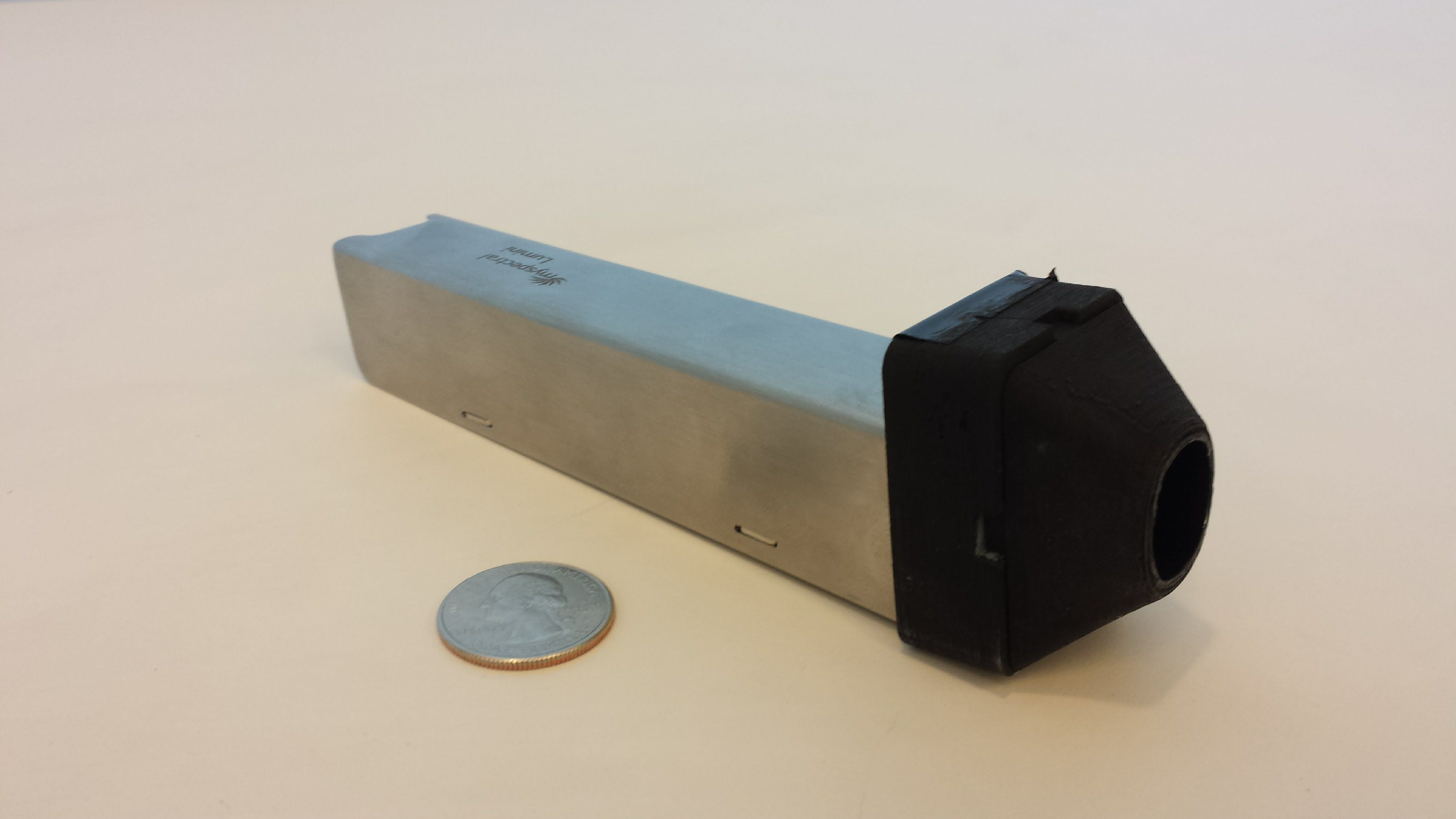}
\includegraphics[width=0.23\textwidth, trim={5cm 2cm 3cm 2cm}, clip]{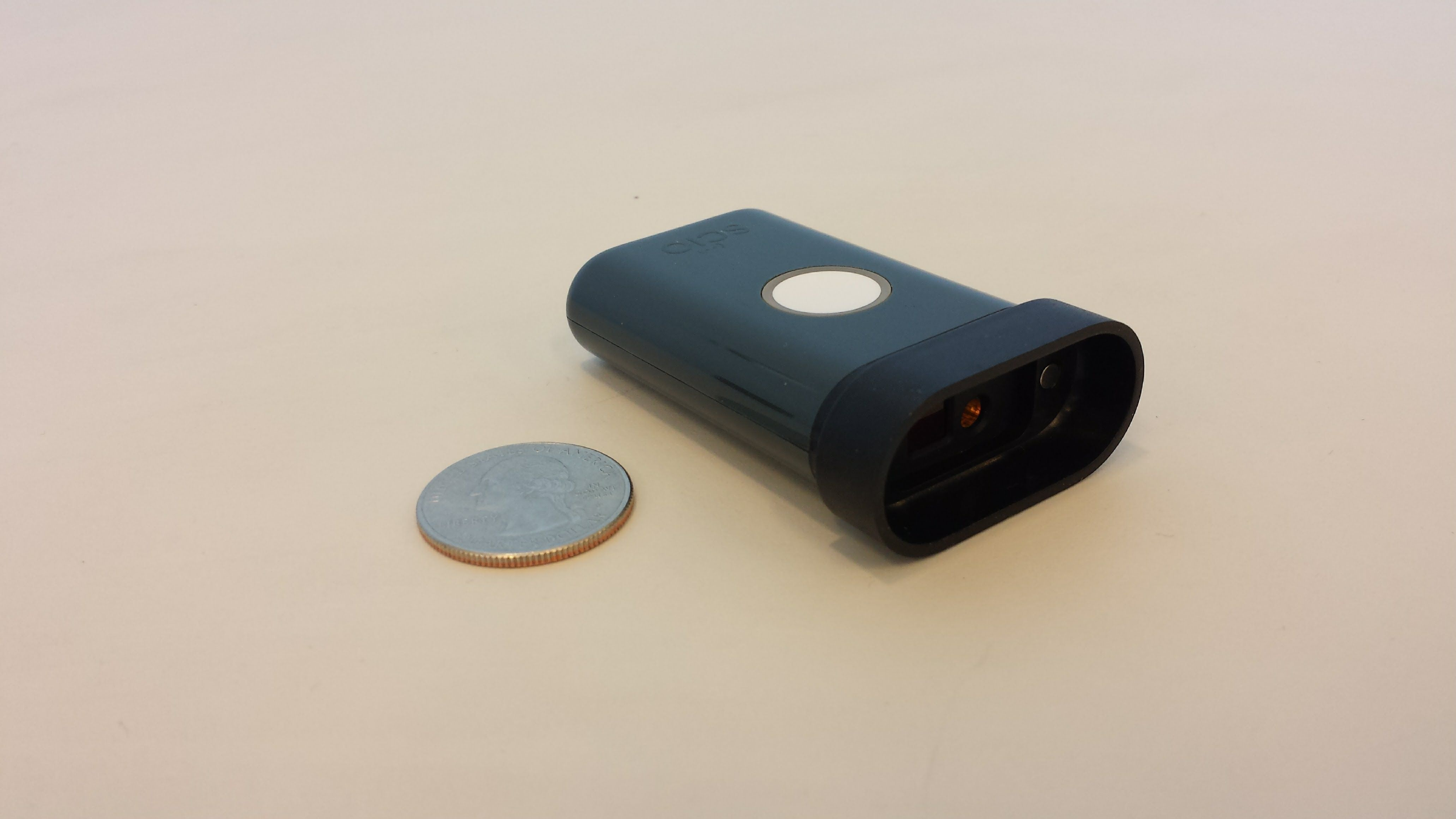}
\caption{\label{fig:sensors}(Left) Lumini sensor. (Right) SCiO sensor. A quarter is provided for sizing. We use black ambient light covers on both sensors to block out environmental light from adding noise into measurements.}
\vspace{-0.2cm}
\end{figure}

We compare two commercially available handheld micro spectrometers, both of which could be held by or attached to a robot's end effector. The first sensor, Lumini ONE\footnote{Lumini ONE: \url{http://myspectral.com}}, seen in Fig.~\ref{fig:sensors} (left), has four illumination sources and dimensions of 18.5 $\times$ 2.4 $\times$ 2.2~cm.
The spectrometer aperture has a wavelength measuring range of $\lambda =$ 317 to 856~nm. This spectrometer covers the entire visible light spectrum as well as parts of the near-ultraviolet and near-infrared spectra.

Fig.~\ref{fig:sensors} (right) depicts the second spectrometer, SCiO\footnote{SCiO: \url{https://www.consumerphysics.com}}. As discussed in Section~\ref{sec:related_work}, the SCiO has shown success in estimating properties such as chemical composition for both foods and pharmaceuticals~\cite{lee2017nir}. When compared to the Lumini, the SCiO has smaller dimensions, measuring at 5.4 $\times$ 3.64 $\times$ 1.54~cm. In addition, the SCiO operates entirely in the near-infrared spectrum at wavelengths of 740-1,070~nm. 

Both spectrometers have black pigmented covers that surround the sensor aperture and help block out environmental light. These covers also ensure that there is an $\sim$1~cm air gap between the sensor aperture and the object. 

\subsection{1-DoF Robot}

\begin{figure}
\centering
\vspace{6pt}
\includegraphics[width=0.48\textwidth, trim={5cm 1cm 5cm 1cm}, clip]{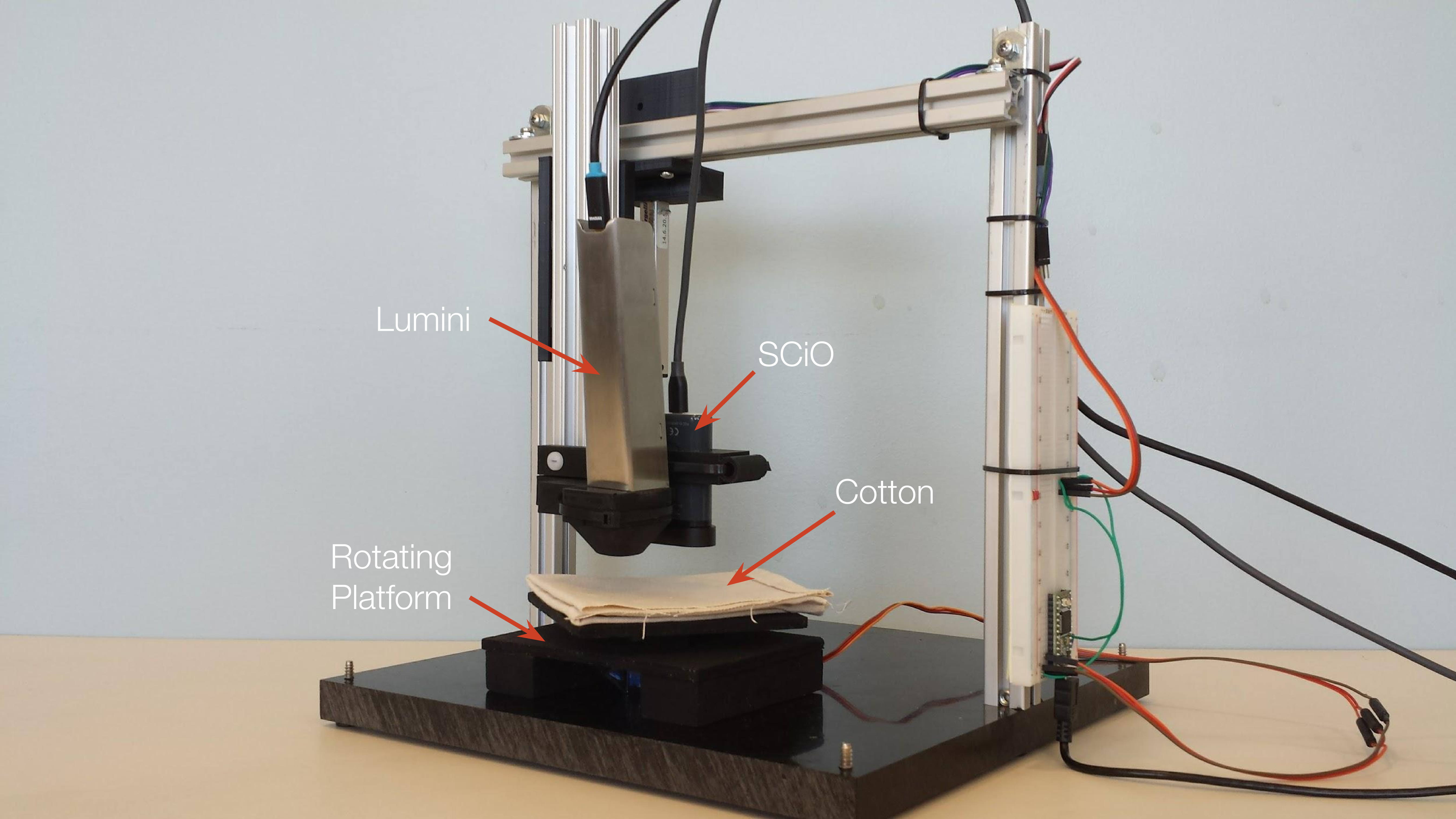}
\caption{\label{fig:robot}The linearly actuated robot we used to collect spectral measurements from materials.}
\vspace{-0.4cm}
\end{figure}

In order to collect sensor measurements, we attached both the Lumini and SCiO sensors onto a 1-DoF robot actuated by a linear servo. This robot and sensor attachment is shown in Fig.~\ref{fig:robot}. We placed material samples on a rotating platform, which would rotate the material sample to a random orientation after each spectral measurement was captured. This rotation helped add variation to the dataset and prevented the spectrometers from continually taking measurements of the same location on an object. We painted the platform with a dark black pigment\footnote{Black 2.0: \url{http://stuartsemple.com}} to absorb stray light that reached the platform, which can occur with translucent or small objects. 

There are several benefits to collecting samples on a controlled platform. The 1-DoF motions promote faster rates of data collection and ensure consistent and repeatable interactions with objects. Furthermore, these idealized spectral measurements allow us to characterize the achievable material recognition performance when these sensors operate in best-case scenarios. In Section~\ref{sec:generalizationpr2} we explore how a model trained on these idealized measurements can generalize to spectral measurements when a mobile manipulator interacts with everyday objects commonly found in the home.

\begin{figure*}
\centering
\vspace{6pt}
\includegraphics[width=0.19\textwidth, trim={2cm 1cm 2cm 1cm}, clip]{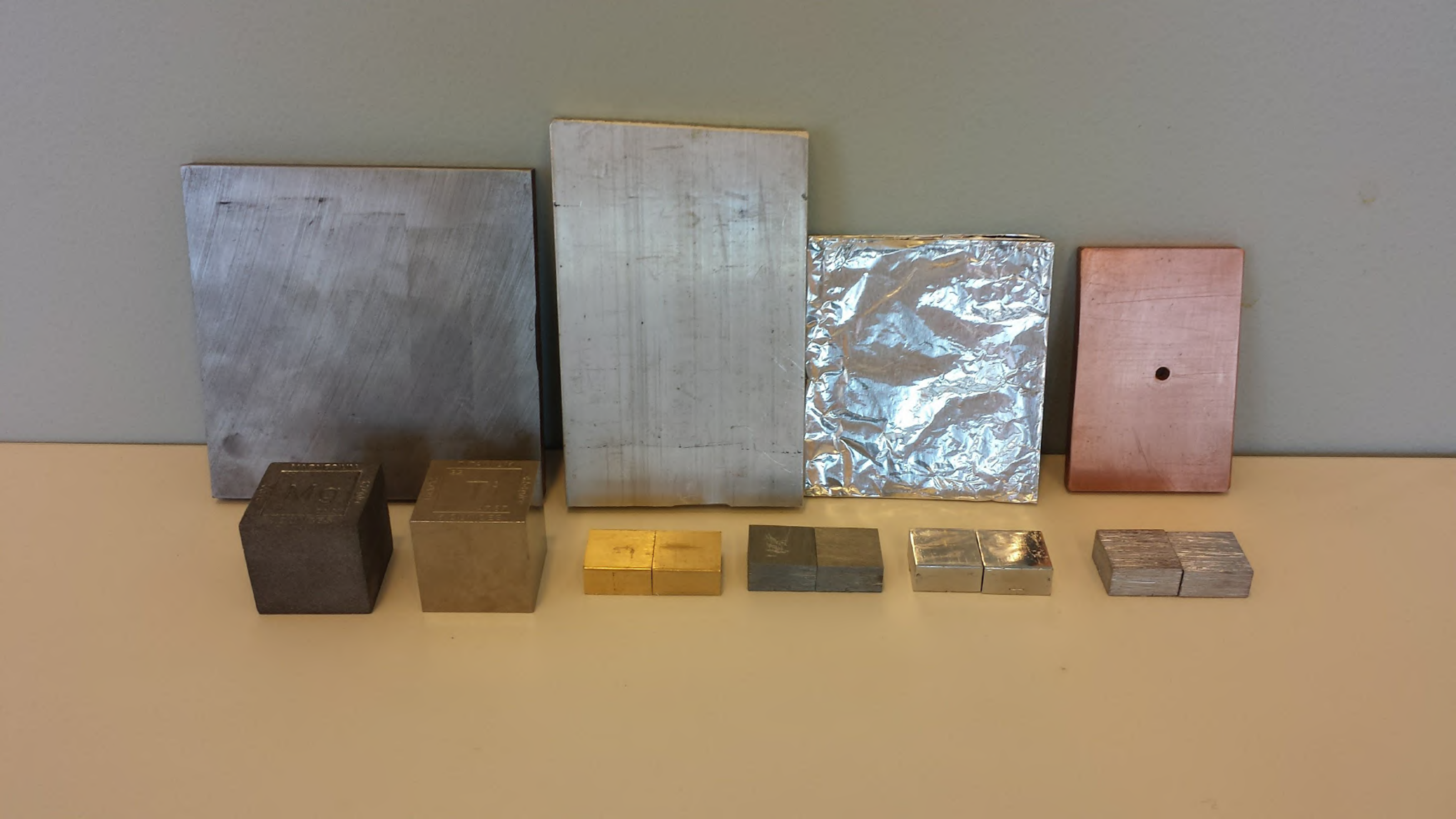}
\includegraphics[width=0.19\textwidth, trim={2cm 1cm 2cm 1cm}, clip]{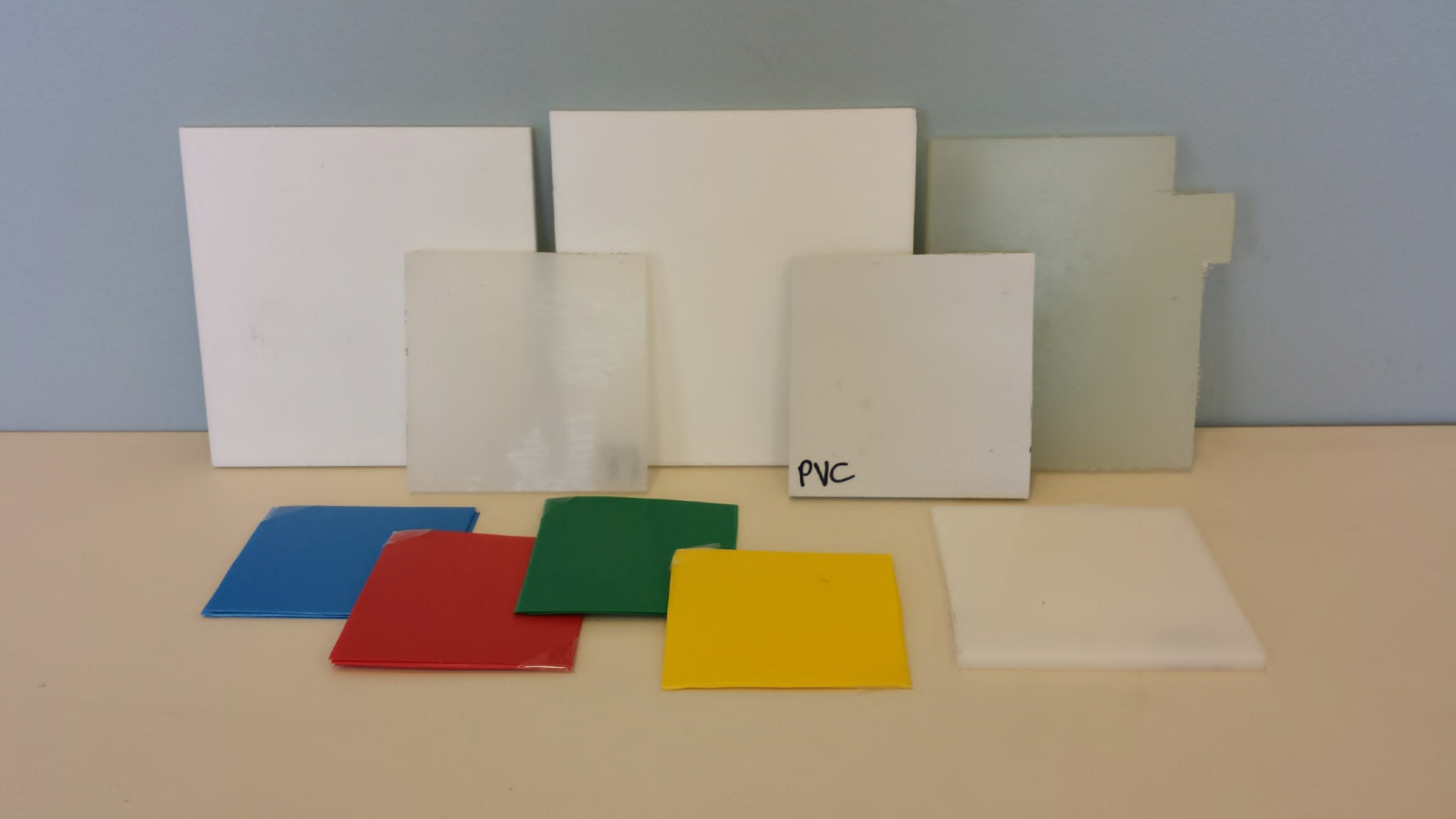}
\includegraphics[width=0.19\textwidth, trim={2cm 1cm 2cm 1cm}, clip]{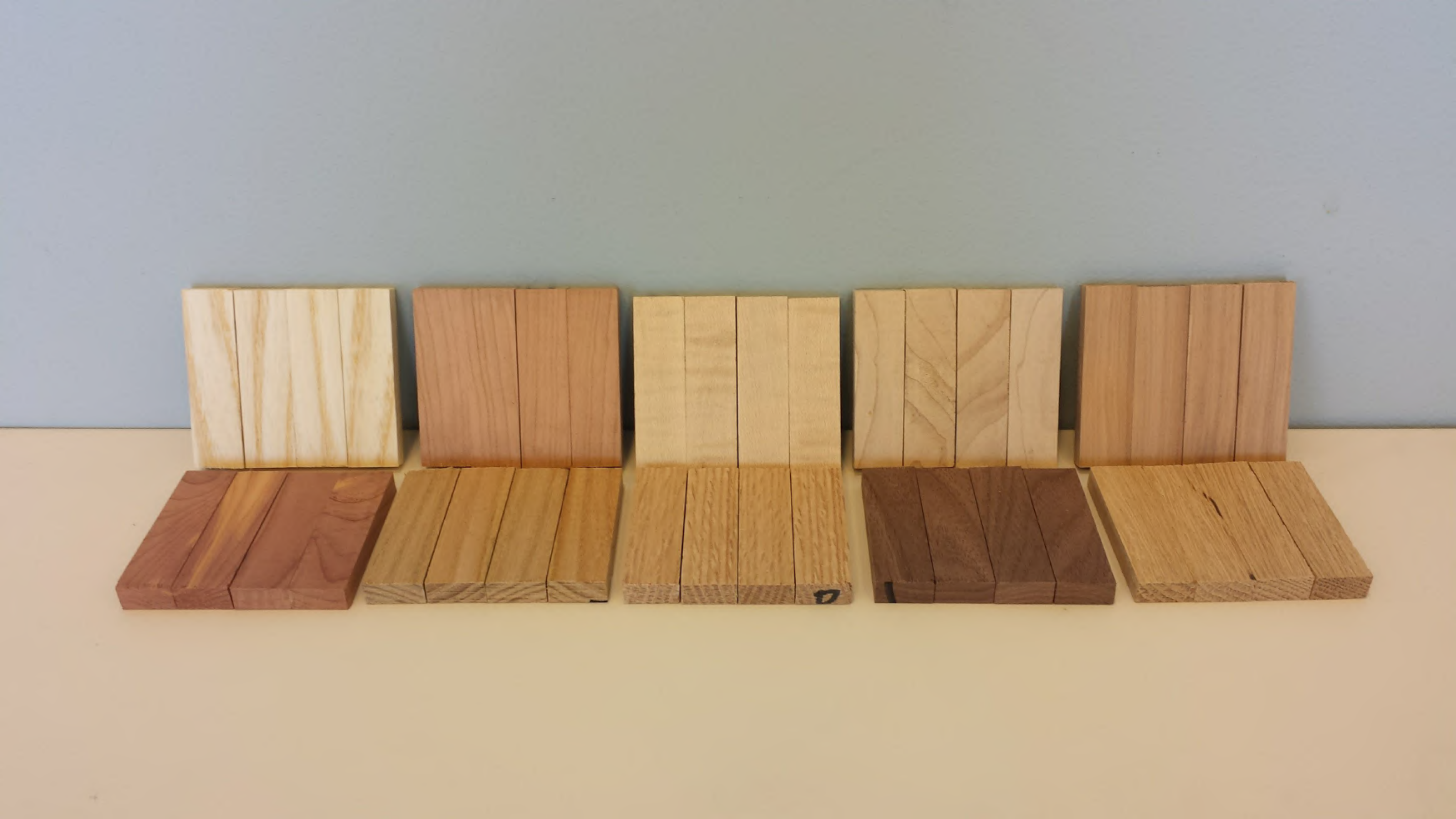}
\includegraphics[width=0.19\textwidth, trim={2cm 2cm 2cm 0cm}, clip]{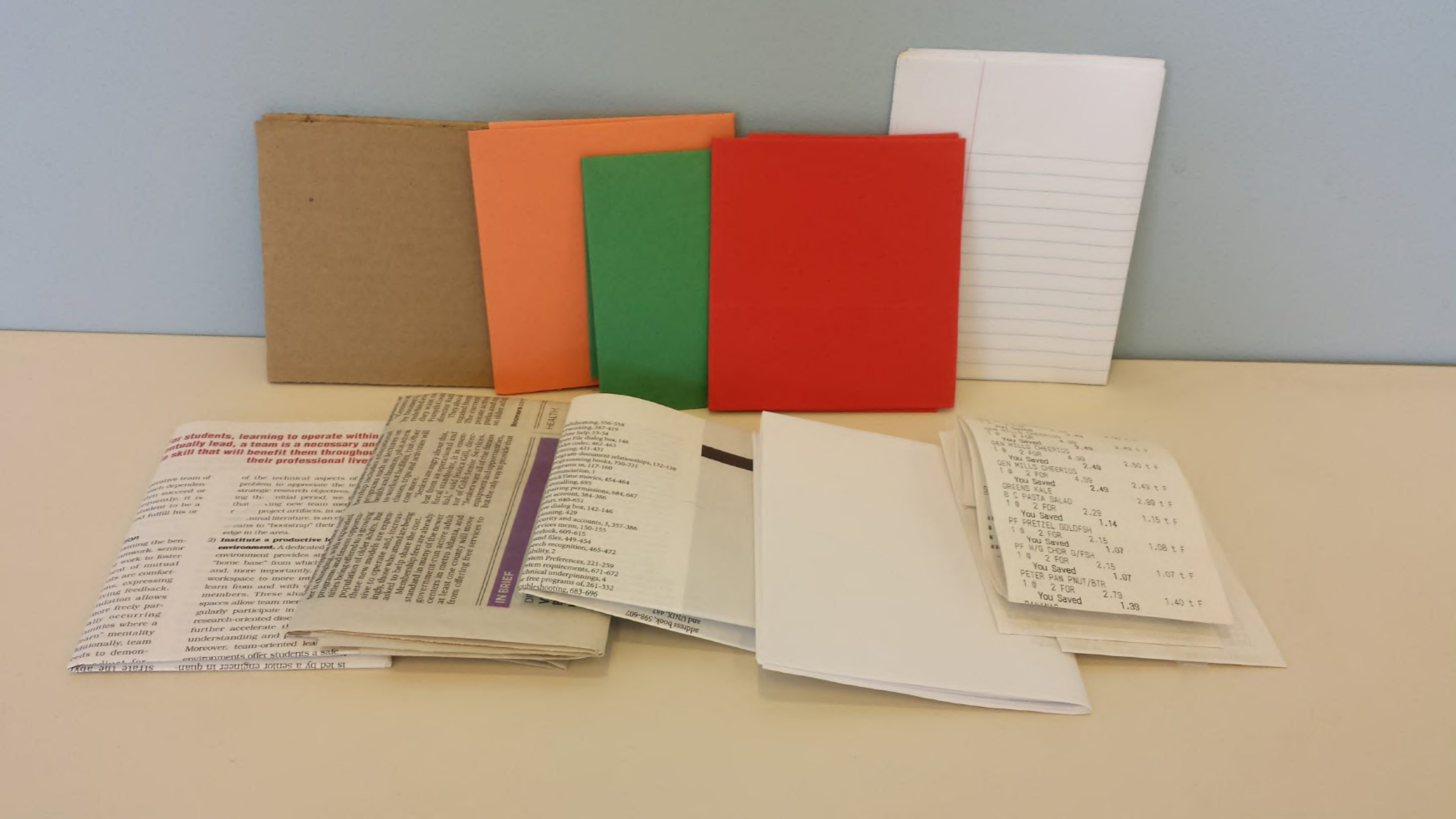}
\includegraphics[width=0.19\textwidth, trim={2cm 2cm 2cm 0cm}, clip]{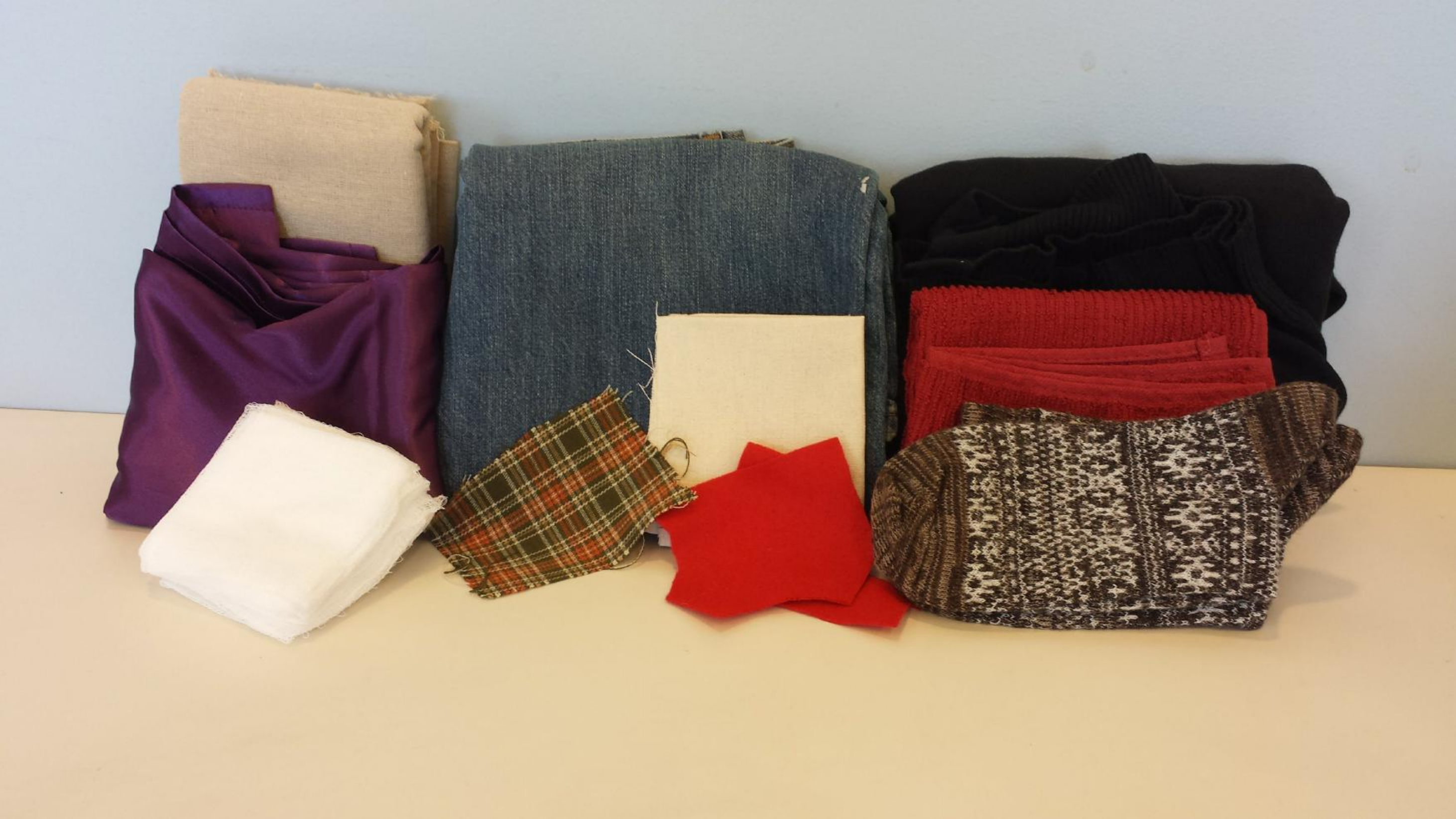}
\caption{\label{fig:imagewall}Each of the five material categories with 10 objects per category. From left to right: metal, plastic, wood, paper, and fabric.}
\vspace{-0.4cm}
\end{figure*}

\begin{figure}
\centering
\includegraphics[width=0.48\textwidth, trim={1cm 18cm 1cm 1.5cm}, clip]{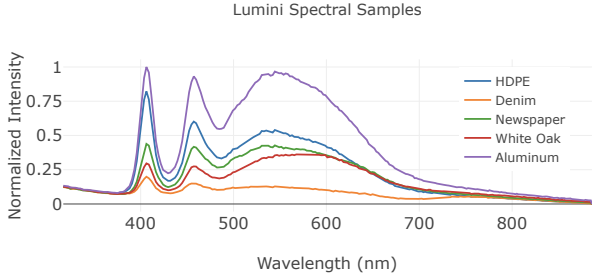}
\caption{\label{fig:luminisignals}Sample spectral signals captured from the Lumini sensor.}
\vspace{-0.3cm}
\end{figure}

\begin{figure}
\centering
\includegraphics[width=0.48\textwidth, trim={1cm 18cm 1cm 1.5cm}, clip]{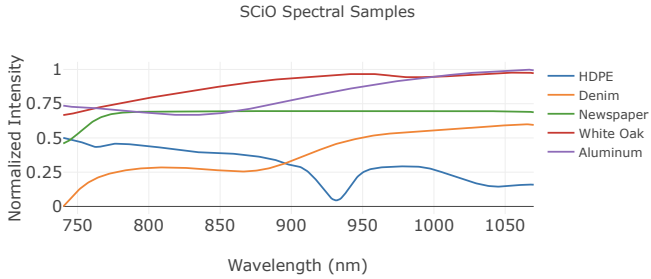}
\caption{\label{fig:sciosignals}Sample spectral signals captured from the SCiO sensor.}
\vspace{-0.4cm}
\end{figure}

\subsection{Spectral Measurements of Materials (SMM50) Dataset}
\label{sec:dataset}



We have released a dataset consisting of 10,000 spectral measurements captured from the 50 objects shown in Fig.~\ref{fig:imagewall}. These objects span five materials categories---metal, plastic, wood, paper, and fabric---with 10 objects per material. A list of all 50 objects can be found in Section~\ref{sec:generalization}. We collected 100 measurements per object for both the Lumini and SCiO sensors, totaling 5,000 observations per spectrometer.

To capture a measurement, the linear actuator moved downwards to a predefined position such that the black covers on both sensors made contact with an object resting on the rotating platform. During these trials, sensors make contact with objects to obtain idealized spectral measurements for characterizing the best-case material recognition performance. However, in Section~\ref{sec:generalizationpr2}, we demonstrate that a PR2 can use these spectrometers to estimate the materials of everyday objects, even when contact does not occur.
We note that spectroscopy is efficient in comparison to many existing haptic sensors used for material estimation. For example, we used an exposure time of 0.5 seconds for the Lumini, meaning contact was only necessary for half a second. We observed that increasing the exposure time increases the intensity of spectral measurements, but also increases noise in the signal. We found an exposure time of 0.5 seconds resulted in a reasonable trade-off between signal intensity and noise. Unlike the Lumini, the SCiO's exposure time is not accessible, but empirically we found it to be $\sim$1 second. Once the spectral measurements were captured, the linear actuator pulled both sensors upwards just enough for the platform to rotate the material sample. 

\begin{figure}
\centering
\includegraphics[width=0.48\textwidth, trim={1cm 19cm 1cm 1.5cm}, clip]{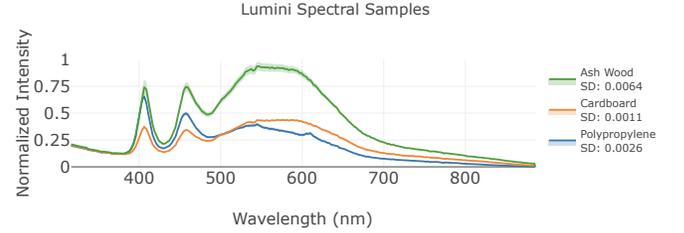}
\caption{\label{fig:lowvariance_lumini}The average over all 100 Lumini spectral samples from the polypropylene, cardboard, and ash wood objects. Background shading represents one standard deviation and demonstrates the similarity between consecutive measurements from the same object. The average standard deviation (SD) across all wavelengths is also provided for each object.}
\end{figure}

\begin{figure}
\centering
\includegraphics[width=0.48\textwidth, trim={1cm 19cm 1cm 1.5cm}, clip]{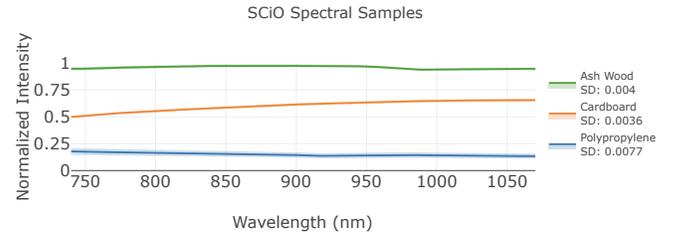}
\caption{\label{fig:lowvariance_scio}The average over 100 SCiO spectral measurements from polypropylene, cardboard, and ash wood. Background shading represents one standard deviation and average standard deviation (SD) is provided for each object.}
\vspace{-0.4cm}
\end{figure}

Fig.~\ref{fig:luminisignals} and Fig.~\ref{fig:sciosignals} show measurements from each material category captured by both the Lumini and SCiO sensors, respectively. Interestingly, we found that there is similarity in consecutive measurements taken from the same object.
For example, Fig.~\ref{fig:lowvariance_lumini} displays the average over all 100 Lumini measurements from the polypropylene, cardboard, and ash wood objects, along with standard deviations for each object when samples are normalized within [0, 1]. Although the spectrometers take measurements at different locations across an object's surface, the resulting spectral responses remain similar with low standard deviations. Fig.~\ref{fig:lowvariance_scio} displays the mean and standard deviation for all 100 SCiO measurements for the same three objects.
In Section~\ref{sec:classification}, we further discuss how this similarity enabled a trained model to recognize the materials of all 50 objects with over 99\% accuracy.
Given the similarity in consecutive spectral samples, we note that future works in material recognition with spectral data may benefit more from a larger array of objects within each material category, rather than additional spectral samples per object.


\section{EVALUATION}

In the following sections, we describe several experiments that explore the use of spectral measurements for material classification. We measured signals from 50 objects across 5 material categories using both the Lumini and SCiO spectrometers, as described in Section~\ref{sec:spectrometer}. The dimensionality of a SCiO measurement is 331 (740~nm to 1070~nm wavelengths), whereas the dimensionality of a Lumini sample is 288 (317~nm to 856~nm with $\sim$1.9~nm between samples). Before learning, we computed the difference quotient (discrete first order derivative) and normalized all measurements, a technique commonly employed in other works that analyze spectral samples~\cite{strother2009nir}. For Lumini measurements, we implemented a fifth order forward-backward Butterworth filter to smooth the signal prior to taking the difference quotient.

\begin{figure}
\centering
\vspace{6pt}
\includegraphics[width=0.48\textwidth, trim={0cm 0cm 0cm 0cm}, clip]{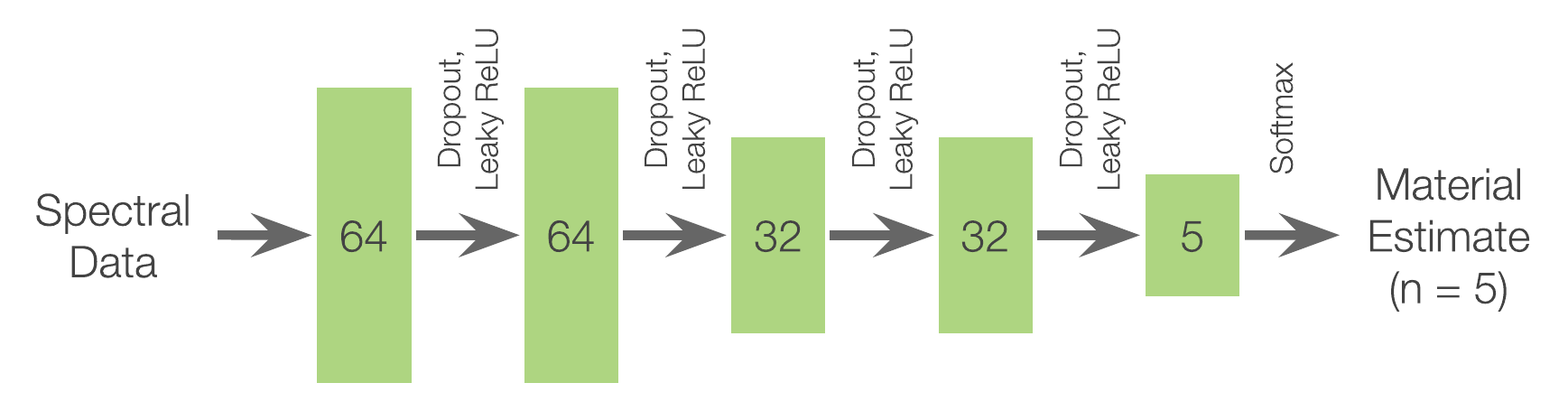}
\vspace{-0.2cm}
\caption{\label{fig:nn}Neural network architecture. We apply a dropout of 0.25 and leaky ReLU activation after each layer with softmax on the final layer output.}
\vspace{-0.4cm}
\end{figure}

To infer material properties, we train a fully-connected neural network, with the architecture in Fig.~\ref{fig:nn}.
We train two models, one that takes SCiO spectral samples (331 dimensional vector) as input, and another that takes Lumini samples (288 dimensional vector) as input. Our models, trained in Keras with TensorFlow, consist of two 64 node layers, followed by two 32 node layers, with a final five node output layer and a softmax activation. We apply a dropout of 0.25 and a leaky ReLU activation after each layer.
We train our models for 300 epochs with a batch size of 32. We use the Adam optimizer with $\beta_1=$~0.9, $\beta_2=$~0.999, and a learning rate of 0.0005.

\subsection{Material Classification}
\label{sec:classification}

Similar to prior works in material recognition, we evaluated how our learned model can recognize new measurements from objects already found in the training set via stratified 5-fold cross-validation. The SMM50 dataset consists of 5,000 spectral measurements from each of the two spectrometers (Lumini and SCiO). Each fold had 1,000 spectral samples with 20 randomly selected samples from each of the 50 objects (200 measurements per material category). In addition to cross-validation, we also evaluated how classification performance varies as we decrease the amount of training data. Specifically, for each training set of four folds (4,000 measurements, 800 per material), we threw away training samples from each of the 50 objects, yet we continued to evaluate the model on all 1,000 samples from the testing fold. Training on 1 sample per object amounts to training on only 50 of the 4,000 measurements. 

Table~\ref{table:t0_lumini} summarizes our results when we performed 5-fold cross-validation on spectral data from both the Lumini and SCiO. Using the Lumini sensor with a 0.5 second exposure time, we achieved an accuracy of 93.8\% with 10 training samples per object (100 samples per material) and 99.2\% accuracy with 40 samples per object (400 per material). In related work with haptic signals, researchers achieved 84\% accuracy with 0.5 seconds of temperature data and 500 samples per material~\cite{bhattacharjee2015material}. Similarly, prior work in semi-supervised material classification with everyday objects has achieved 81.4\% accuracy with 120 samples per material and 4 seconds of force and temperature data~\cite{erickson2017semi}. However, note that both prior works relied on haptic sensors with objects and materials that differ from those used in our work and hence we cannot provide a direct comparison of results.

On the other hand, using spectral measurements from the SCiO, we achieved a classification accuracy of 99.9\%. One feature of the SCiO sensor is that we needed only one spectral measurement of an object in order to recognize the object's material with 94.6\% accuracy.
This indicates that there exists a similarity between SCiO measurements of the same object, whereas measurements from different objects are easily separable, as visually depicted in Fig.~\ref{fig:lowvariance_scio}.
In further support of this notion, we find that a simple linear SVM trained on SCiO measurements achieved 98.8\% classification accuracy with 10 training samples per object, and 99.9\% accuracy with all 80 samples per object.

\subsection{Generalizing to New Objects via Leave-One-Object-Out}
\label{sec:generalization}

In the prior section, we observed that spectral measurements can enable a robot to estimate the material of objects found in the training set with high success. However, in real-world environments, there will often be times when a robot interacts with a new object that it has not been trained to recognize. In this and the following sections, we explore how well our model can use spectral measurements to classify objects that are not found in the training data. Here, we assess generalization across all 50 available objects using leave-one-object-out cross-validation. This technique consists of training a model on 49 objects (4,900 measurements, 100 from each object) and evaluating classification performance on the 100 measurements from the one remaining object. We then repeat this process for each object and report results as the average accuracy over all 50 training and test sets.

With leave-one-object-out cross-validation, our model achieved an accuracy of 68.7\% on Lumini measurements and an accuracy of 79.1\% with measurements from the SCiO sensor. Similar to results in Section~\ref{sec:classification}, near-infrared spectral samples with the SCiO increased recognition accuracy.
In comparison, a linear SVM underperformed when generalizing to new objects, achieving an accuracy of only 58.5\% and 72.9\% for Lumini and SCiO measurements, respectively.

\begin{table}
\centering
\vspace{6pt}
\caption{\label{table:t0_lumini}Material recognition with Lumini and SCiO samples. Results are presented as accuracies computed via cross-validation.}
\begin{tabular}{ccccc} \toprule
    & \multicolumn{4}{c}{\# of training samples per object} \\ \cmidrule{2-5}
     & 1 & 10 & 40 & 80 (100\%) \\ \midrule\midrule
    Lumini & 65.1 & 93.8 & 99.2 & 99.8 \\
    SCiO & \textbf{94.6} & \textbf{99.3} & \textbf{99.9} & \textbf{99.9} \\
	\bottomrule
\end{tabular}
\vspace{-0.4cm}
\end{table}

\begin{figure}
\centering
\includegraphics[width=0.48\textwidth, trim={0cm 0cm 0cm 0cm}, clip]{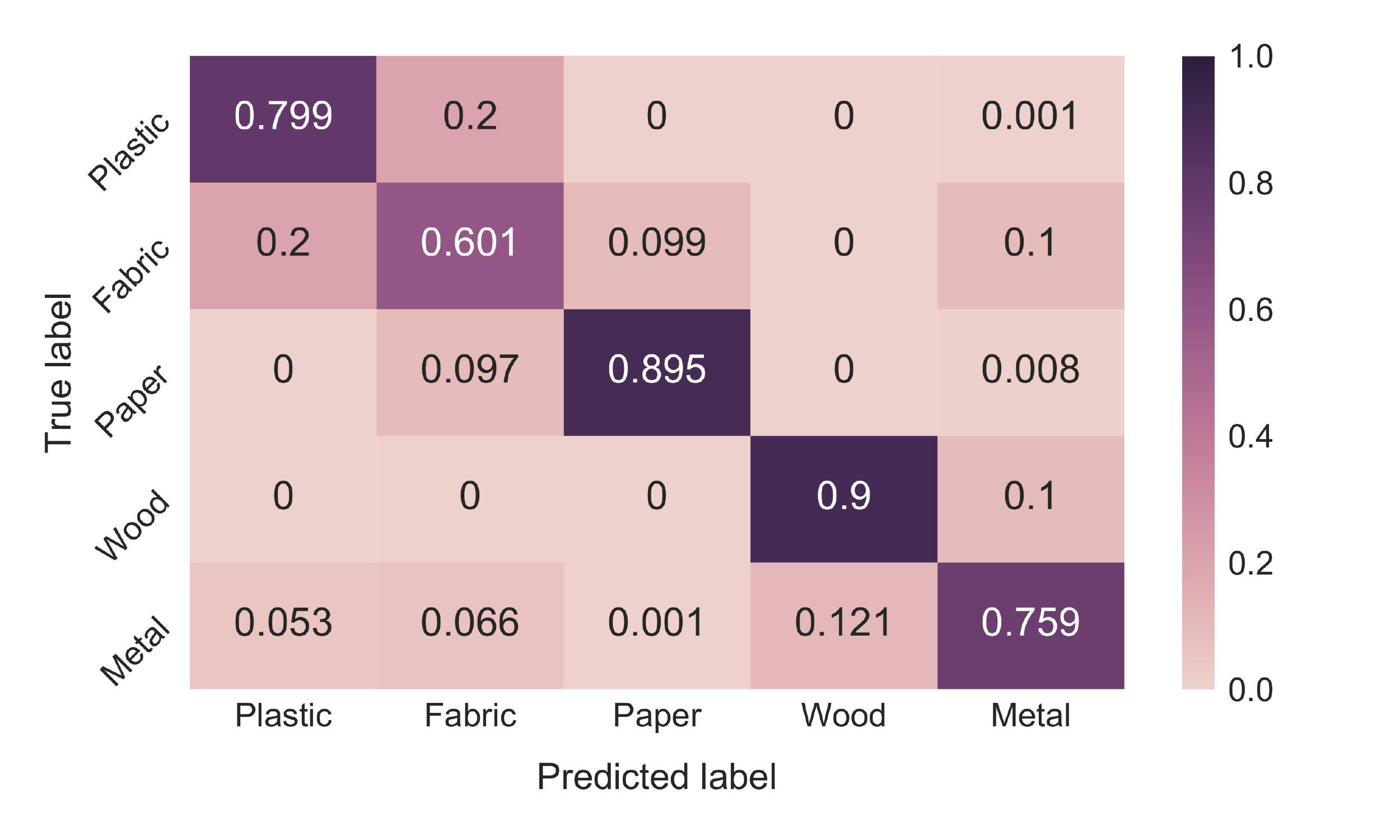}
\vspace{-0.4cm}
\caption{\label{fig:confusionmatrix}Confusion matrix for classifications of SCiO measurements using leave-one-object-out cross-validation. Results are presented as accuracies.}
\vspace{-0.4cm}
\end{figure}

\begin{figure}
\centering
\includegraphics[width=0.48\textwidth, trim={0cm 0cm 0cm 0cm}, clip]{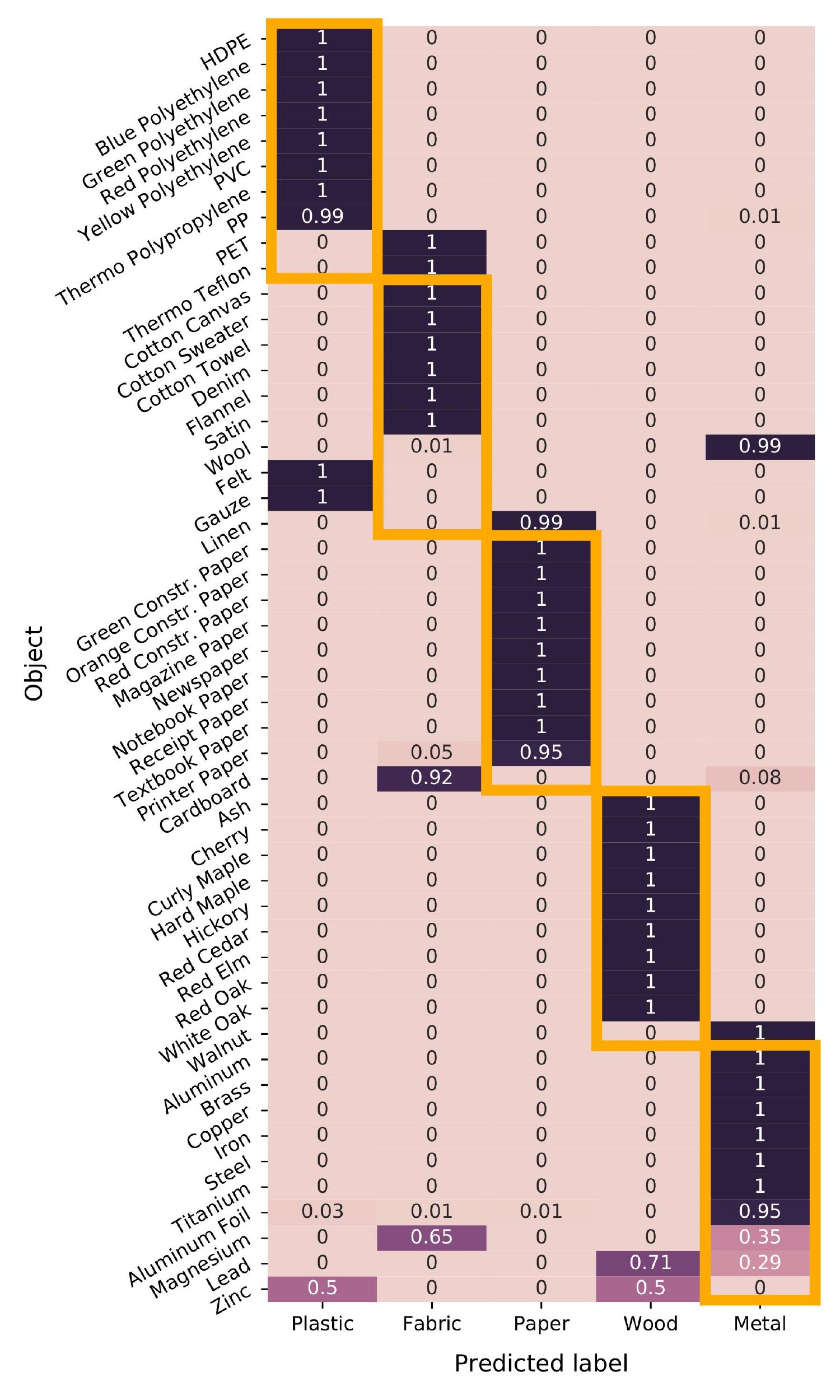}
\vspace{-0.4cm}
\caption{\label{fig:all50objects}Material classification results for each object when testing the generalization capabilities of the model on SCiO measurements with leave-one-object-out cross-validation. Results are presented as accuracies. Correct material labels are highlighted in orange. Objects within each category are sorted in descending order by classification performance.}
\vspace{-0.4cm}
\end{figure}

In addition to overall accuracy, we also explored where our neural network model performed well, and what materials are difficult for generalization. Fig.~\ref{fig:confusionmatrix} displays a confusion matrix of how our model classified SCiO measurements from each material category. We observed that fabric objects posed the most challenges for generalization, whereas our model generalized well to both paper, and wood objects.
Interestingly, some measurements of paper were misclassified as fabric, yet none were misclassified as wood. Fig.~\ref{fig:all50objects} shows a list of all 50 objects along with a detailed breakout of how our model classified all 5,000 SCiO measurements from these objects during leave-one-object-out cross-validation. Regions outlined in orange represent the correct material label for a set of objects. We notice that when the model misclassifies a left out object, it often misclassifies all 100 samples.

\begin{figure}
\centering
\vspace{6pt}
\includegraphics[width=0.48\textwidth, trim={1cm 17.75cm 1cm 3.25cm}, clip]{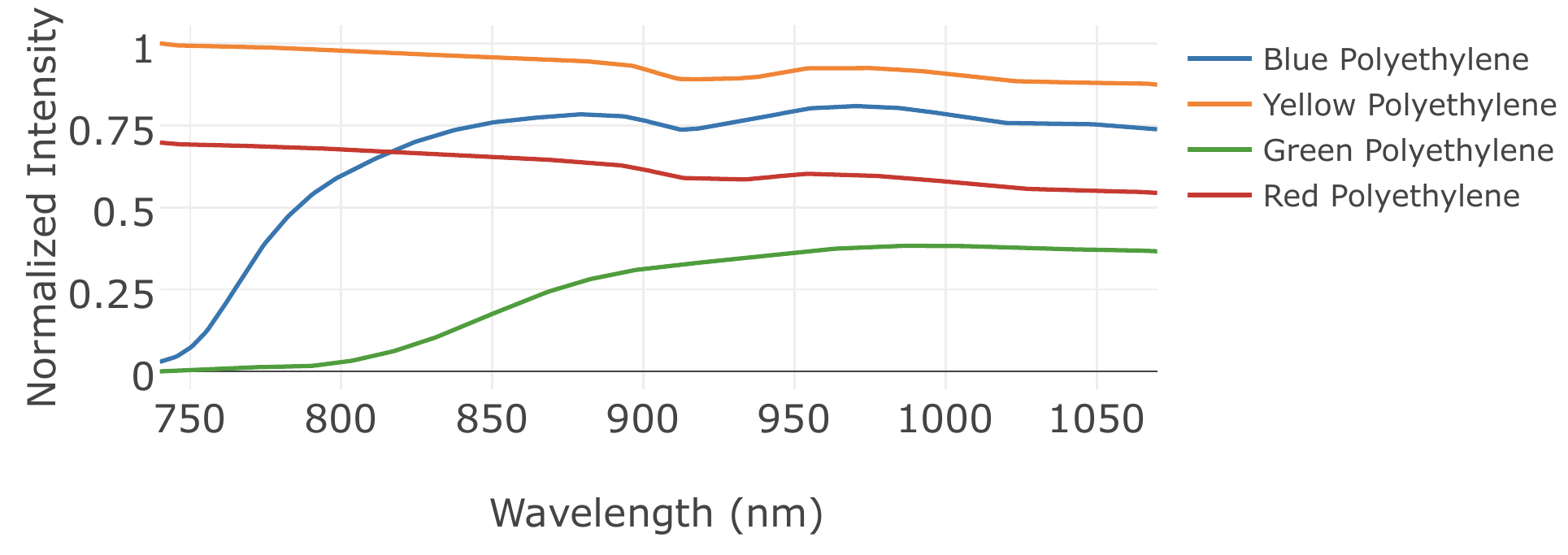}
\vspace{-0.2cm}
\caption{\label{fig:colorgeneralization}SCiO samples from four color variants of polyethylene plastic.}
\end{figure}

Furthermore, we investigate the impact a material's color has on spectral measurements. Given that the Lumini operates over the entire visible light spectrum, it may not be surprising that the Lumini observes different measurements for different color variants of the same material. However, it is unclear if the SCiO sensor---which operates solely in the near-infrared spectrum---will also be affected by varying colors of a material. In Fig.~\ref{fig:colorgeneralization}, we plot sample SCiO measurements captured from four different color variations (blue, yellow, green, and red) of the same polyethylene plastic. The appearance of these four colored plastics can be found in Fig.~\ref{fig:imagewall}. We observe that the specific coloring of an object \textit{can} impact near-infrared measurements from the SCiO spectrometer, yet these color variations did not impact the generalization capabilities of our model. Despite the vibrant color differences between the four plastics, our model was able to recognize all four as plastic with 100\% accuracy during generalization, as seen in Fig.~\ref{fig:all50objects}.

\subsection{Generalization with More Training Objects}

At the end of Section~\ref{sec:dataset}, we presented a notion that training models on a larger variety of objects may provide more benefit than collecting and training on a larger number of measurements per object. In this section, we test and provide evidence towards that notion. 

Previously, we explored the generalization capabilities of our model, which resulted in a leave-one-object-out cross-validation accuracy of 79.1\% on SCiO measurements. We can extend this generalization approach by evaluating how performance scales as we vary the number of training objects. Specifically, we continued to use leave-one-object-out cross-validation, but rather than train on 49 objects at each iteration, we trained on only $n\in\{1,\ldots,10\}$ objects from each material, while still testing on all 50 objects individually. 

\begin{figure}
\centering
\includegraphics[width=0.48\textwidth, trim={1cm 17cm 1cm 3.25cm}, clip]{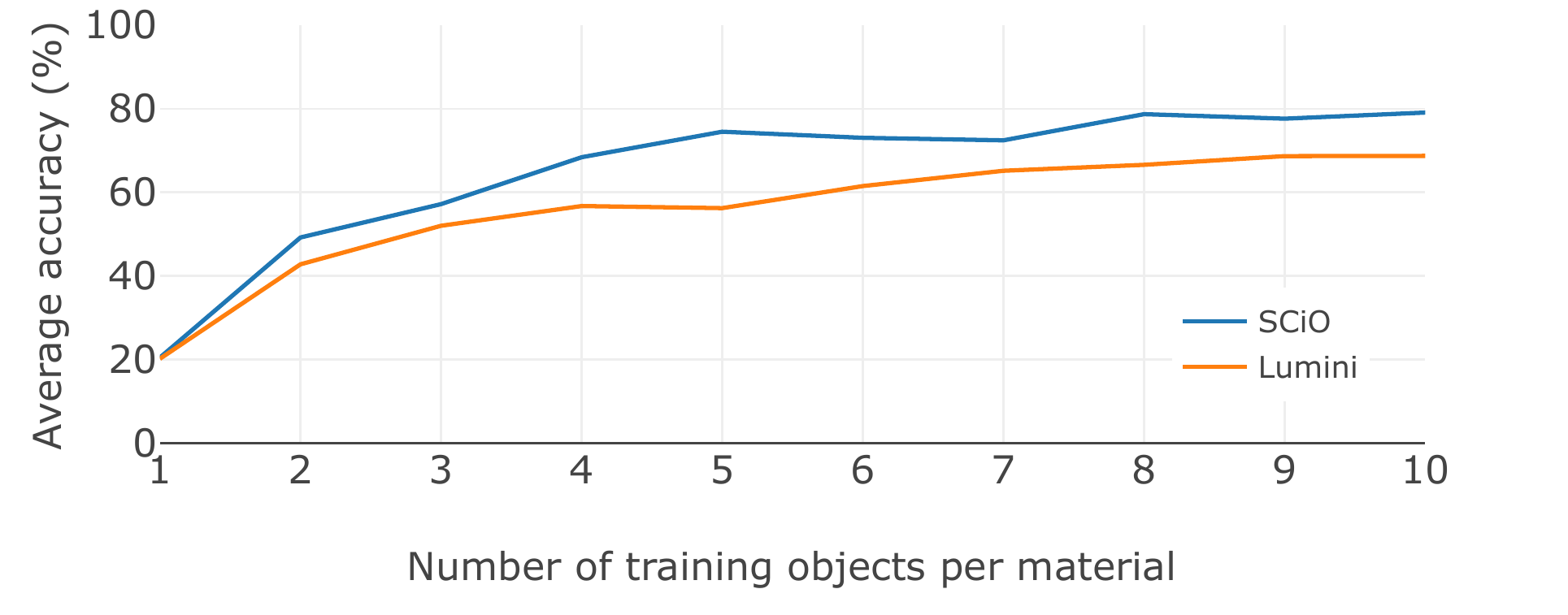}
\vspace{-0.2cm}
\caption{\label{fig:looo_luminiscio}Leave-one-object-out performance as we increase the number of training objects.}
\vspace{-0.4cm}
\end{figure}

Fig.~\ref{fig:looo_luminiscio} shows how generalization performance varies as we increased the number of objects used during training. Results are computed as the average over 50 trained models (one for each left out object). For both Lumini and SCiO measurements, we observe that performance gradually improves as we increase the number of training objects in each material category. These results suggest that the classification and generalization performance of models trained on spectral data will continue to improve as these models are trained on increasing numbers of objects from each material category.

\subsection{Generalizing to Everyday Objects with a PR2}
\label{sec:generalizationpr2}

In prior sections, we explored how spectroscopy can be used to estimate the materials of objects given an idealized robotic platform. Here, we demonstrate the use of spectroscopy with a mobile manipulator that interacts with objects found in household environments. For this, we employed a Willow Garage PR2 robot that used both the Lumini and SCiO spectrometers to estimate the materials of 25 everyday objects, 5 per material category, as seen in Fig.~\ref{fig:pr2_objects}. These objects were selected to be significantly different from objects found in the SMM50 dataset that was used for model training. For example, several of the everyday metal objects were made of stainless steel or tin, neither of which are found in the training data.

In order to estimate the material of an object, the robot grasped one of the two spectrometers with its parallel jaw gripper. We then placed objects at a known location on a table in front of the robot. Objects rested on the table and were not adhered to the table's surface. For each object, the robot performed 10 consecutive interactions consisting of reaching towards a random location along the outer surface of the object and capturing a spectral measurement. For objects that had sufficient height, such as bottles and bowls, the robot's end effector followed a horizontal linear motion until the spectrometer reached the random target position near the object's surface. For objects that laid flat on the table, such as books, plates, and fabrics, the robot's end effector started above the object and performed a vertical motion downwards towards the object. Prior to an interaction, the robot also randomly rotated the spectrometer $\pm$30 degrees, along the central axis of the gripper. These interactions can be seen in greater detail in the supplementary video.

\begin{figure}
\centering
\includegraphics[width=0.48\textwidth, trim={0cm 0cm 0cm 0cm}, clip]{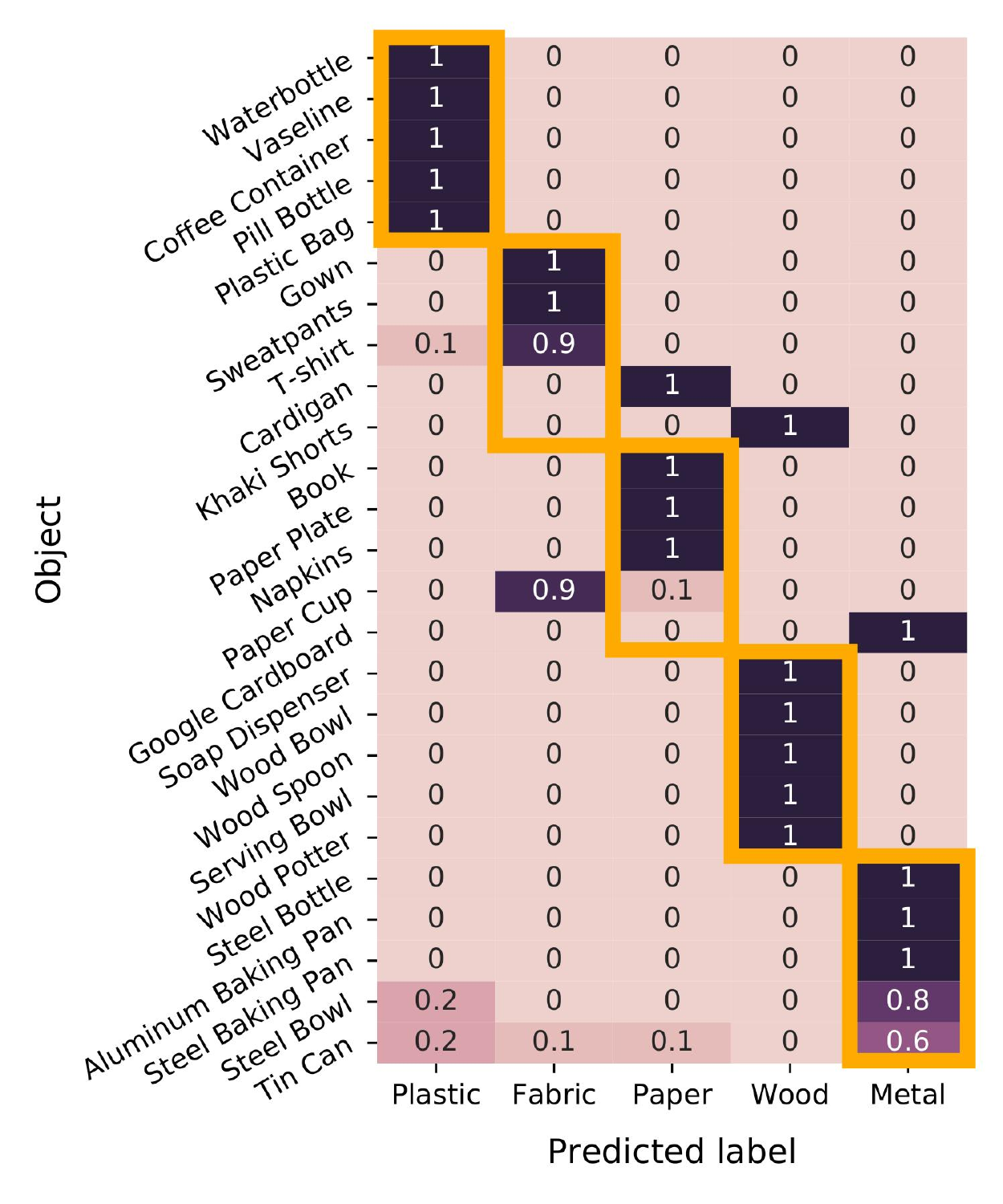}
\vspace{-0.4cm}
\caption{\label{fig:pr2_confusionmatrix}Material recognition performance for each object that the PR2 interacted with using the SCiO spectrometer. Results are presented as accuracies. Correct material labels are highlighted in orange. Objects within each category are sorted in descending order by classification performance.}
\vspace{-0.4cm}
\end{figure}

Using a model trained on all 5,000 SCiO measurements from the SMM50 dataset, the PR2 was able to estimate the materials of these 25 everyday objects with 81.6\% accuracy across all 250 interactions (10 per object). Notably, this performance matches that observed during leave-one-object-out cross-validation in Section~\ref{sec:generalization}. In addition, these results provide evidence that, when used to estimate an object's material properties, near-infrared spectral measurements are \textit{not} significantly impacted by varying shapes of objects or by stray environmental light from indoor fluorescent bulbs reaching the sensor aperture. Fig.~\ref{fig:pr2_confusionmatrix} depicts how the 10 spectral measurements from each object were classified. Using the SCiO spectrometer, the robot was able to perfectly recognize all plastic and wood objects. Yet, similar to leave-one-object-out results from Fig.~\ref{fig:all50objects}, we observe that fabrics remained more difficult for our model to recognize. 


Despite the SCiO's success when generalizing to everyday objects, we note that our model struggled to generalize measurements from the Lumini sensor. Specifically, when using a model trained on 5,000 Lumini measurements, the PR2 achieved an accuracy of only 34.8\% when estimating the materials of these everyday objects across the 250 interactions. We find that measurements from the Lumini are noticeably impacted by environmental lighting. In contrast to the linearly actuated robot with flat material samples, these interactions with everyday objects allowed for a substantial amount of visible light to reach the sensor's aperture. This is especially true for objects with curved surface contours, such as bowls and cups, which reflected environmental light into the aperture.

We note that there remains significant room for future improvement when generalizing material classification to everyday objects that a robot may interact with. For example, transparent materials and objects with dark black pigments are challenging for spectrometers to sense accurately. Combining spectroscopy with other visual or haptic modalities may alleviate these issues and improve overall recognition performance.

\section{DISCUSSION}

Overall, we observe that visible light spectral measurements from the Lumini are negatively impacted by environmental lighting, which commonly occurs during interactions with everyday objects. Despite this, we find that near-infrared samples from the SCiO are robust to both changes in objects shapes and robotic platforms. Models trained on near-infrared measurements from an idealized robotic scenario showed promising results when used by a PR2 to recognize the materials of everyday objects. These trained models are small and can estimate the material label for a new spectral measurement in under 50 milliseconds using only the PR2's on-board CPUs.

It is worth noting that some household objects can be composed of multiple material types. In these cases, inferring multiple material labels may be more appropriate, and this presents an opportunity for future study.

In prior work, a generative adversarial network (GAN) achieved a material recognition accuracy of 75.1\% using leave-one-object-out cross-validation on temperature, force, and vibration signals collected from a robot that interacted with 72 household objects~\cite{erickson2017semi}. With the SCiO, our PR2 and trained model recognized the materials of 25 household objects with 81.6\% accuracy, which suggests that the use of spectroscopy for material recognition may remain competitive with haptic sensors, even when generalizing to new objects found throughout the home.

Although direct contact between a sensor's black cover and an object can help block out environmental light, we found that contact is not always necessary. During many of the PR2's interactions with objects, there was up to a 1~cm gap between the sensor's black cover and an object, in part due to approximate placing of objects. Yet, with the SCiO, this lack of contact with everyday objects did not worsen generalization when comparing the PR2 study results (81.6\% accuracy) with leave-one-object-out cross validation results (79.1\% accuracy) from the idealized robotic platform. It is also worth noting that other classification methods might further improve performance. The deep networks we used outperformed other classifiers, including linear SVMs, throughout our study.

Finally, while both the SCiO and Lumini are small enough to be held in a robot's end effector, it is feasible that these spectrometers could be integrated directly into a robot's fingertip. The actual spectrometer sensor inside these devices is $\sim$1 cubic cm in size, which has enabled the SCiO to be integrated into some commercial smartphones already. The small size of these spectrometers also presents an opportunity for future work to explore embedding both spectroscopy and haptic sensing into the same robot end effector.

\section{CONCLUSION}



In this work, we presented how robots can leverage spectral data to infer the materials of objects. We compared two commercially available handheld micro spectrometers, Lumini and SCiO, both of which cover a different region of the electromagnetic spectrum. We have also presented a new dataset of 10,000 spectral measurements from these two sensors collected on a robotic platform that interacted with 50 objects across five material categories.

We observed a similarity in consecutive spectral measurements from the same object, and this enabled our models to achieve highly accurate material classification performance with few training samples. 

When using near-infrared spectral measurements to estimate the materials of new objects not in the training set, our model remained competitive with prior results from haptic sensors.
When used to estimate the material of an object, we found that near-infrared spectral measurements from the SCiO consistently outperformed measurements from the Lumini spectrometer, which relies on the visible light spectrum.

Furthermore, we show that a PR2, mobile manipulator, can use this spectroscopy approach to infer the materials of 25 everyday objects, such as cups, bowls, and garments. Unlike measurements from the Lumini sensor, we found that near-infrared spectral measurements from the SCiO were robust to changes in both object shapes and robotic platforms.


Through this work, we have demonstrated that spectroscopy presents a reliable and effective way for robots to infer the material properties of everyday household objects.

\bibliographystyle{IEEEtran}
\bibliography{bibliography}

\end{document}